\pdfoutput=1

\documentclass[11pt,a4paper]{article}
\usepackage[hyperref]{acl}
\usepackage{times}
\usepackage{inconsolata}
\usepackage{latexsym}
\usepackage{danudefs}
\usepackage{algorithmic}
\usepackage{algorithm}
\usepackage{color}
\usepackage{booktabs}
\usepackage{xspace}
\usepackage{url}
\usepackage{braket}
\usepackage[T1]{fontenc}
\usepackage[utf8]{inputenc}
\usepackage{graphicx}
\usepackage{microtype}
\usepackage{tabularx}
\usepackage{multicol}
\usepackage{multirow}
\usepackage{makecell}
\usepackage{bm}
\usepackage{subcaption}
\usepackage{array}  
\usepackage{caption}
\usepackage{booktabs}
\usepackage{xcolor}

\usepackage[english]{babel}
\usepackage{hyperref}
\addto\captionsenglish{%
}
\addto\extrasenglish{%
}

\usepackage{acronym}

\usepackage{etoolbox}
\makeatletter
\newif\if@in@acrolist
\AtBeginEnvironment{acronym}{\@in@acrolisttrue}
\newrobustcmd{\LU}[2]{\if@in@acrolist#1\else#2\fi}

\newcommand{\ACF}[1]{{\@in@acrolisttrue\acf{#1}}}
\makeatother

\acrodef{NLP}[NLP]{Natural Language Processing}
\acrodef{IR}[IR]{Information Retrieval}
\acrodef{QA}[QA]{Question Answering}
\acrodef{SoTA}[SoTA]{state-of-the-art}

\acrodef{MLM}[MLM]{Masked Language Model}
\acrodefplural{MLM}[MLMs]{Masked Language Models}
\acrodef{LLM}[LLM]{Large Language Model}
\acrodefplural{LLM}[LLMs]{Large Language Models}
\acrodef{MLP}[MLP]{Multi-layer Perception}
\acrodef{FFN}[FFN]{Feed Forward Network}

\acrodef{ICL}{\LU{I}{i}n-cotext \LU{L}{l}earning}
\acrodef{ITML}[ITML]{Information-Theoretic Metric Learning}
\acrodef{SDML}[SDML]{Semantic Distance Metric Learning}
\acrodef{CASE}[CASE]{Condition-Aware Sentence Embeddings}
\acrodefplural{CASE}[CASEs]{Condition-Aware Sentence Embeddings}
\acrodef{PCA}[PCA]{Principal Component Analysis}
\acrodef{ICA}[ICA]{Independent Component Analysis}
\acrodef{IPC}[IPC]{Isotropy approximated by Principle Components}

\acrodef{SCD}{\LU{S}{s}emantic \LU{C}{c}hange \LU{D}{d}etection}
\acrodef{WiC}{Word-in-Context}
\acrodef{C-STS}[C-STS]{Conditional Semantic Textual Similarity}
\acrodef{STS}[STS]{Semantic Textual Similarity}
\acrodef{KGC}[KGC]{Knowledge Graph Completion}

\acrodef{MTEB}[MTEB]{Massive Text Embedding Benchmark}

\acrodef{NV}[NV]{NV-Embed-v2}
\acrodef{SFR}[SFR]{SFR-Embedding-Mistral}
\acrodef{GTE}[GTE]{gte-Qwen2-7B-instruct}
\acrodef{E5}[E5]{Multilingual-E5-large-instruct}
\acrodef{SimCSE-large}[SimCSE\_large]{sup-simcse-roberta-large}
\acrodef{SimCSE-base}[SimCSE\_base]{sup-simcse-bert-base-uncased}

\usepackage{todonotes}

\makeatletter
\if@todonotes@disabled

\else

\fi
\makeatother

\title{\emph{CASE} -- Condition-Aware Sentence Embeddings for \\Conditional Semantic Textual Similarity Measurement}

\author{
Gaifan Zhang\textsuperscript{1} \quad
Yi Zhou\textsuperscript{2} \quad
Danushka Bollegala\textsuperscript{1,3} \\
\textsuperscript{1} University of Liverpool \quad
\textsuperscript{2} Cardiff University \quad
\textsuperscript{3} Amazon \\
\texttt{\{sggzhan8,danushka\}@liverpool.ac.uk, zhouy131@cardiff.ac.uk}
}

\date{}

\begin{document}

\maketitle

\begin{abstract}
    The meaning conveyed by a sentence often depends on the context in which it appears. 
    Despite the progress of sentence embedding methods, it remains unclear as how to best modify a sentence embedding conditioned on its context.
    To address this problem, we propose \ac{CASE}, an efficient and accurate method to create an embedding for a sentence under a given condition. 
    First, \ac{CASE} creates an embedding for the condition using a \ac{LLM} encoder, where the sentence influences the attention scores computed for the tokens in the condition during pooling.
    Next, a supervised method is learnt to align the \ac{LLM}-based text embeddings with the \ac{C-STS} task.
    We find that subtracting the condition embedding consistently improves the \ac{C-STS} performance of \ac{LLM}-based text embeddings by improving the isotropy of the embedding space.
    Moreover, our supervised projection method significantly improves the performance of \ac{LLM}-based embeddings despite requiring a small number of embedding dimensions.\footnote{code: \url{https://github.com/LivNLP/CASE}}
\end{abstract}

\section{Introduction}
\label{sec:intro}
\begin{figure}
    \centering
    \includegraphics[width=1.0\linewidth]{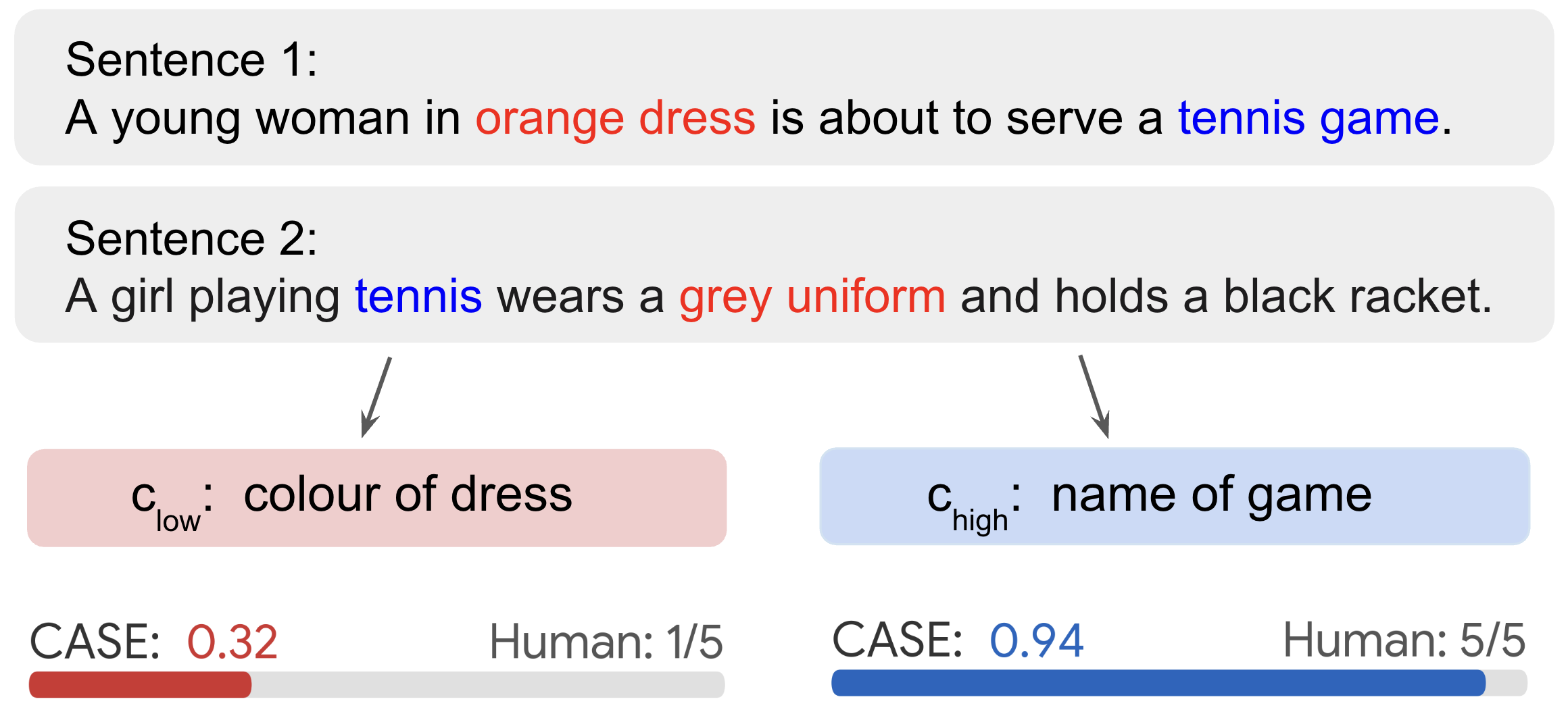}
    \caption{The two conditions focus on different information described in the two sentences. Human annotators rate the two sentences 1--5, indicating a high-level (5) of semantic textual similarity under $c_{\rm high}$ than $c_{\rm low}$ (1). Our proposed condition-aware sentence embedding (CASE) method reports similarity scores that are well-aligned with the human similarity ratings.}
    \label{fig:intro}
\end{figure}

Measuring \ac{STS} between sentences is a fundamental task in NLP~\cite{majumder2016semantic}.
It is also important in training the sentence encoders~\cite{sentence-bert}.
Recent approaches, such as contrastive learning, use semantic similarity as an objective to improve the quality of sentence embeddings~\cite{Gao:2021c}.
However, measuring sentence similarity is a complex task, which depends on the aspects being considered in the sentences being compared.
To address this problem, \citet{deshpande-etal-2023-c} proposed the \ac{C-STS} task along with a human-annotated dataset.
They designed the \ac{C-STS} task and the dataset by assigning different conditions (semantic aspects) to each pair of sentences ($s_1$, $s_2$), a condition of high similarity $c_{\rm high}$ and a condition of low similarity $c_{\rm low}$, resulting in different similarity ratings for the same pair of sentences.
This reduces subjectivity and ambiguity in the measurement of similarity between two sentences.
As shown in~\autoref{fig:intro}, human annotators are required to assign different similarity scores under different conditions, rather than a single score as in the traditional STS task.
Many real-world applications can be seen as \ac{C-STS} tasks such as ranking a set of documents retrieved for the same query in \ac{IR}~\cite{Manning:2008}, comparing two answers to the same question in \ac{QA}~\cite{Risch:2021}, or measuring the strength of a semantic relation between two entities in \ac{KGC}~\cite{Hyper-CL,Lin:2024}.

\begin{figure*}[t!]
    \centering
    \includegraphics[width=1.0\linewidth]{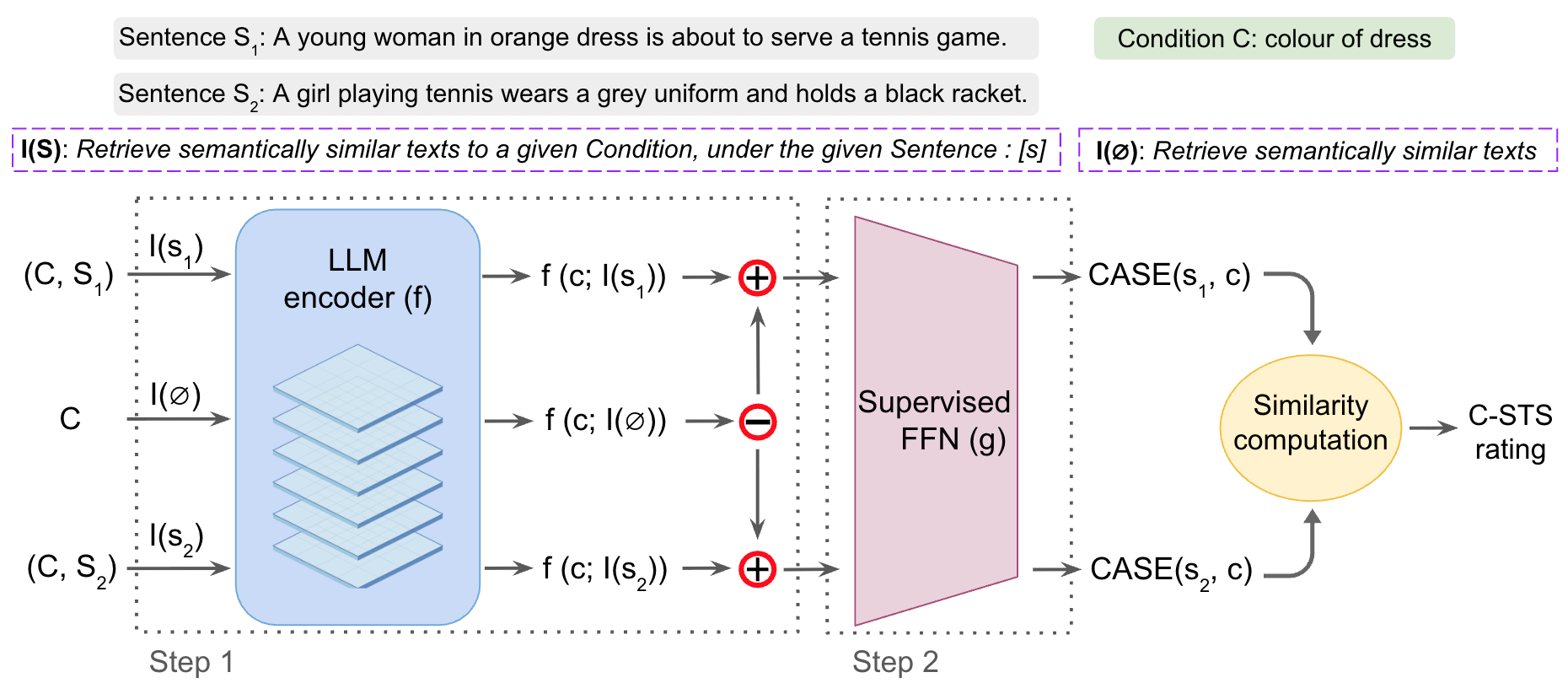}
    \caption{Overview of \ac{CASE}. An LLM is prompted with $I(s)$ to obtain two separate embeddings $f(c;I(s_1))$ and $f(c;I(s_2))$ for the same condition $c$ for the two sentences $s_1$ and $s_2$. The unconditional embedding $f(c;I(\emptyset))$ is then computed using the prompt $I(\emptyset)$ and subtracted from each of those conditional embeddings. Finally, the embeddings are projected to a lower-dimensional space using a supervised \ac{FFN} and their cosine similarity is computed as the \ac{C-STS} rating.}
    \label{fig:overview}
\end{figure*}

We propose \textbf{\ac{CASE}} (\textbf{C}ondition-\textbf{A}ware \textbf{S}entence \textbf{E}mbeddings), an efficient two-step method to combine \ac{LLM}-based sentence encoders and a lightweight supervised projection, to create the embeddings required for the \ac{C-STS} task (see \autoref{fig:overview}).
In the first step, we prompt the \ac{LLM} encoders to create the initial embeddings, with an instruction based on conditional semantic similarity to encode the condition given a sentence.
Here, the sentence is not encoded explicitly but instead influences the attention scores during token pooling.
Compared to \ac{MLM}-based encoders that have been used in prior work on \ac{C-STS}, \ac{LLM}-based encoders can be used as accurate embedding models by direct prompting~\cite{tao2025llmseffectiveembeddingmodels}.
They benefit from large-scale datasets and complex architectures with billions of parameters, excelling for various downstream tasks, as demonstrated by the performance on the MTEB leaderboard~\cite{muennighoff-etal-2023-mteb}.
However, the best way to use \acp{LLM} for \ac{C-STS} remains elusive, as reported by \citet{Lin:2024}, who showed that decoder-only \acp{LLM} often underperform \ac{MLM}-based embeddings in \ac{C-STS} benchmarks.
Interestingly, we find that our prompt structure ``\emph{retrieve [condition]}, given [sentence]'' counter-intuitively performs better than ``\emph{retrieve [sentence]}, given [condition]''.
Furthermore, a post-processing step of subtracting the embedding of the condition significantly improves the performance and isotropy of the embedding space (as shown in \autoref{sec:exp}).

Previous \ac{C-STS} methods fine-tune sentence encoders~\cite{SEAVER, L-CSTS, Hyper-CL, Lin:2024}, which is computationally expensive, particularly for \ac{LLM} encoders due to their significantly larger number of parameters (>1B), even with parameter-efficient fine-tuning methods.
Moreover, \ac{LLM} encoders produce high-dimensional embeddings (e.g. 4096 for \texttt{NV-embed-v2}~\cite{lee2024nv}) compared to \ac{MLM} encoders (e.g. 768 for SimCSE~\cite{Gao:2021c}), posing challenges when storing and computing inner-products between embeddings.
To address these issues, the second step of \ac{CASE} trains a lightweight supervised \acf{FFN} (<5M) in a bi-encoder setting to fine-tune the embeddings in a lower-dimensional space.
Qualitative analysis shows that \ac{CASE} increases the similarity between a sentence and the information emphasised by a given condition, while decreasing the same for information irrelevant to the condition, as expected.

Our \ac{CASE} method makes the following contributions:
\begin{itemize}
    \setlength\itemsep{0em} 
    \setlength\parskip{0em}
    \setlength\parsep{0em}
    \item It uses the superior semantic capabilities of \ac{LLM}-based embedding models.
    \item It avoids fine-tuning or re-training existing encoders by using a lightweight \ac{FFN}.
    \item It maintains high performance even with an 8$\times$ reduction in dimensionality (e.g., from 4096 to 512).
\end{itemize}

\section{Related Work}
\label{sec:related}

\citet{deshpande-etal-2023-c} proposed the \ac{C-STS} task and created a dataset containing 18,908 instances where the semantic similarity between two sentences $s_1$ and $s_2$ is rated under two conditions $c_{\rm high}$ and $c_{\rm low}$ resulting in, respectively, high vs. low similarity between the two sentences.
Moreover, they proposed cross-, bi- and tri-encoder baselines.
Cross-encoders consider interactions between all tokens in $s_1$, $s_2$ and $c$, which are computationally expensive for large-scale comparison due to a lack of pre-computed conditional embeddings.
Tri-encoders separately encode $s_1$, $s_2$ and $c$, and then apply late interactions between the condition's and each sentence's embeddings, which are also complex and report suboptimal performance.
Bi-encoders overcome those limitations by creating a condition-aware embedding for each sentence and computing \ac{C-STS} efficiently as their inner-product.

\citet{L-CSTS} found multiple issues in the dataset created by \citet{deshpande-etal-2023-c}, such as poorly defined conditions and inconsistent similarity ratings for more than half of the dataset.
To address this, \citet{L-CSTS} re-annotated the validation split from the original \ac{C-STS} dataset.
Moreover, they proposed a QA-based approach for measuring \ac{C-STS} by first converting conditions into questions, and then using GPT-3.5 to extract the corresponding answers, which are the compared using their embeddings.
This QA formulation depends on multiple decoupled components, such as converting conditions into questions, requiring a decoder \ac{LLM} to extract answers, and using a separate encoder to generate embeddings, which increases the possibility of propagation of errors between components.

\citet{zhang-etal-2025-annotating} further investigated the dataset created by \citet{deshpande-etal-2023-c} and found issues such as problematic condition statements and inaccurate human annotations.
They proposed an \ac{LLM}-based two-step method to improve the dataset by first refining the condition statements and then re-annotating the ratings, creating a larger scale (14176 instances) dataset of better quality, which we use in our experiments as training data.

Alternative architectures to the tri-encoder and cross-encoder have been explored to capture conditional semantics.
\citet{Hyper-CL} proposed Hyper-CL, a tri-encoder using contrastively learnt hypernetwork~\cite{HyperNets} to selectively project the embeddings of $s_1$ and $s_2$ according to $c$.
Hypernetworks introduce an external parameter set that is three times larger than that of the SimCSE model used to encode each sentence, requiring in an excessively large memory space, which is problematic when processing large sets of sentences.
\citet{Lin:2024} proposed a tri-encoder-based \ac{C-STS} method where they used routers and heavy-light attention~\cite{Ainslie:2023a} to select the relevant tokens to a given condition.
\citet{liu2025conditional} proposed a conditional contrastive learning method for \ac{C-STS}, introducing a weighted contrastive loss with a sample augmentation strategy.
\citet{SEAVER} proposed a cross-encoder approach, which predicts \ac{C-STS} scores, without creating conditional embeddings.
They used a token re-weighting strategy by computing two cross-attention matrices between $(s_1, s_2)$ and $c$, which are subsequently used to compute the correlations for the sentence or condition tokens.
Although the aforementioned methods improve the performance of cross-encoder and tri
-encoder-based \ac{C-STS} measurement, they still underperform bi-encoders.

\citet{li-li-2024-bellm} proposed BeLLM, which uses backward dependency to enhance \acp{LLM} by learning embeddings from uni- to bi-directional attention layers.
They fine-tuned the LLaMA2-7B model with their method on the \ac{C-STS} dataset, outperforming previously proposed fine-tuned models.
In particular, they designed a prompt template for this task: ``\emph{Given the context [condition], summarise the sentence [sentence] in one word:}''.
They then extracted embeddings from the hidden states of the generated output and used these embeddings to measure sentence similarity.
Their results showed that larger model sizes and backward dependencies contribute to the fine-tuning effects.
In contrast to our \ac{CASE}, which uses \ac{LLM} encoders, they use decoder-only \acp{LLM}.
Moreover, we do not require token generation and therefore are not restricted to single-word summarisation.
Additionally, we find that the reversed structure (``\emph{given [sentence], retrieve [condition]}'') performs better.

\citet{yamada2025outoftheboxconditionaltextembeddings} proposed an unsupervised conditional text embedding method that uses a causal \ac{LLM}. 
Specifically, they use the input prompt ``\emph{Express this text [sentence] in one word in terms of [condition]},'' and compute the cosine similarity of the last token's hidden state as the \ac{C-STS} score. 
They further validated that instruction-tuned models outperform their non-instruction-tuned counterparts. 
Similarly, our \ac{CASE} leverages prompt-based \ac{LLM} encoders to generate conditional embeddings, yet achieves performance comparable to the best supervised bi-encoder settings in a low-dimensional space.

\section{Condition-Aware Sentence Embeddings}
\label{sec:CASE}


An overview of our proposed method is shown in \autoref{fig:overview}, which consists of two-steps.
In the first step (\autoref{sec:LLM-embeddings}), we create two separate embeddings for the condition considering each of the two sentences, one at a time, in the instruction prompt shown to an \ac{LLM}.
Note that it is the condition that is being encoded and the tokens in the sentence (similar to all other tokens in the instruction) are simply modifying the attention scores computed for the tokens in the condition.
Intuitively, it can be seen as each sentence \emph{filling} some missing information required by the condition.
However, note that the embeddings obtained from \acp{LLM} are not necessarily aligned with the \ac{C-STS} task.
Therefore, in the second step (\autoref{sec:projection}), we learn a projection layer using the training split from the \ac{C-STS} dataset.
Finally, the \ac{C-STS} between two sentences is computed as the cosine similarity between the corresponding projected embeddings.



\subsection{Extracting Embeddings from LLMs}
\label{sec:LLM-embeddings}



Given an \ac{LLM}-based encoder, $f$, we create a $d$-dimensional embedding that captures the semantic relationship between a sentence $s$ and a condition $c$.
There are two ways to formulate the input: 
\begin{enumerate}
\item \textbf{Encode the condition given the sentence}, denoted by $f(c; I(s))$.
Here, $I$ is an instruction template that takes $c$ as an argument.
We use the following prompt template as $I(s)$ --- \emph{Retrieve semantically similar texts to a given Condition, given the Sentence : [s]}, where we substitute $s$ in the placeholder [s].
Next, we provide $c$ as the input text to be encoded by the \ac{LLM} following the instruction $I(s)$.
Finally, the token embeddings of $c$ are aggregated according to one of the pooling methods to create  $f(c; I(s))$ (different pooling methods are discussed in \autoref{sec:app:pooling}).
\item \textbf{Encode the sentence given the condition}, denoted by $f(s; I(c))$.
Here we swap the sentence and the condition in the above formulation, recalling that both $s$ and $c$ are text strings.
In particular, as shown later in our experiments (\autoref{sec:C-STS}), comparing the embedding for $c$ created given $s_1$ and $s_2$ results in better performance on the \ac{C-STS} benchmark for all \ac{LLM} encoders.
For this reason, we keep the best-performing $f(c; I(s))$ in \autoref{fig:overview} and our method for subsequent experiments.
This is because $s_1$ and $s_2$ will also contain information irrelevant to $c$, which will affect the cosine similarity computed between $f(s_1; I(c))$ and $f(s_2; I(c))$. 
On the other hand, the cosine similarity between $f(c; I(s_1))$ and $f(c; I(s_2))$ is a more accurate estimate of \ac{C-STS} between $s_1$ and $s_2$ under $c$ because it is purely based on contextualised representation shifts of $c$ conditioned on $s_1$ and $s_2$ separately.
\end{enumerate}

Condition statements contain their own intrinsic semantics (e.g., the phrase \emph{colour of dress}), which is irrelevant when distinguishing specific semantic values based on sentences (e.g., \emph{orange} vs. \emph{grey}).
To address this problem, we introduce a post-processing step of subtracting the embedding of the condition to reduce the effect of tokens in the condition that are irrelevant to the sentence, thereby consistently improving the accuracy of the condition-aware embeddings.
Specifically, we use the prompt $I(\emptyset)$ \emph{Retrieve semantically similar texts} to obtain the embedding of a condition $c$, and denote this by $f(c; I(\emptyset))$.
As we see later in our experiments, by subtracting $f(c; I(\emptyset))$ from $f(c; I(s))$, we improve the performance on the \ac{C-STS} task and the isotropy of embeddings.
This first step is fully unsupervised and a zero-shot prompt template is used as $I$.


\subsection{Supervised Projection Learning}
\label{sec:projection}

The \ac{LLM} embeddings computed in \autoref{sec:LLM-embeddings} has two main drawbacks.
First, although \acp{LLM} are typically trained on massive text collections and instruction-tuned for diverse tasks~\cite{muennighoff-etal-2023-mteb}, their performance on \ac{C-STS} tasks have been poor~\cite{Lin:2024}.
As seen from our condition-aware prompt template, an \ac{LLM} must be able to separately handle a variable condition statement and a fixed instruction. 
This setup is different from most tasks on which \acp{LLM} are typically trained on, where the instruction remains fixed across all inputs.
Second, relative to \acp{MLM}-based sentence embeddings, \acp{LLM} produce much higher dimensional embeddings, which can be problematic due to their memory requirements (especially when operating on a limited GPU memory) and the computational cost involved in inner-product computations.
In tasks such as dense retrieval, we must compare millions of documents against a query to find the nearest neighbours under strict latency requirements, and low-dimensional embeddings are preferable.

To address the above-mentioned drawbacks, we propose a supervised projection learning method. 
Specifically, we freeze the model parameters of the \ac{LLM} and use a \ac{FFN} layer that takes $f(c; I(s))$ as the input and returns a $k$-dimensional ($k\leq d)$ embedding $g(f(c; I(s); \theta)$, where $\theta$ denotes the learnable parameters of the \ac{FFN}. 
Finally, we define ${\rm CASE}(s,c)$ as the projection of the offset between the conditional and the unconditional embeddings of $c$ under $s$, given by,
\begin{align}
    \label{eq:CASE}
    {\rm CASE}(s,c) = g(f(c; I(s)) - f(c; I(\emptyset)); \theta)
\end{align}
We use the human-annotated similarity ratings $r$ in the \ac{C-STS} train instances $\cD$ to learn $\theta$. 
Specifically, we minimise the squared error between the human ratings and the cosine similarity computed using the corresponding \ac{CASE} as given by \eqref{eq:loss}.
\par\nobreak
{\small
\vspace{-3mm}
\begin{align}
    \label{eq:loss}
    \sum_{(s_1, s_2, c, r) \in \cD} \left( \cos({\rm CASE}(s_1,c), {\rm CASE}(s_2, c)) - r\right)^2
\end{align}
}%
Here, 
$\cos$ denotes the cosine similarity between the projected embeddings.
We use Adam optimiser~\cite{Kingma:2014} to find the optimal $\theta$ that minimises the loss given by \eqref{eq:loss}.
Recall that only the \ac{FFN} parameters are updated during this projection learning step, while keeping the parameters of the \ac{LLM} fixed, which makes it extremely efficient.
Using the learnt projection, we compute the \ac{C-STS} between $s_1$ and $s_2$ under $c$ as the cosine similarity between the embeddings ${\rm CASE}(s_1,c)$ and ${\rm CASE}(s_2, c)$.

\section{Experiments and Results}
\label{sec:exp}


To evaluate the effectiveness of the \ac{LLM}-based and \ac{MLM}-based sentence embeddings as described in \autoref{sec:CASE}, we use six sentence encoders, out of which three are LLM-based: 
\textit{NV-Embed-v2} (4096 dimensional \textbf{\acs{NV}}), 
\textit{SFR-Embedding-Mistral} (4096 dimensional \textbf{\acs{SFR}}), 
\textit{gte-Qwen2-7B-instruct} (3584 dimensional \textbf{\acs{GTE}}) and three are MLM-based: 
\textit{Multilingual-E5-large-instruct} (1024 dimensional \textbf{\acs{E5}}), \textit{sup-simcse-roberta-large} (1024 dimensional \textbf{\acs{SimCSE-large}}), and \textit{sup-simcse-bert-base-uncased} (768 dimensional \textbf{\acs{SimCSE-base}}).
Further details provided in \autoref{sec:app:models}.

\begin{table}[t!]
    \centering
    \small
    \begin{tabular}{l l l}
        \toprule
        \textbf{Model}  & \textbf{sent/cond?} & \textbf{Spear.} \\
        \midrule
        \acs{NV}& sent - c & 16.98 \\
         &  sent & 22.07 \\
         &  \textbf{cond - c} & \textbf{37.61} \\
         &  cond & 33.85 \\
        \midrule
        \acs{SFR}  & sent - c & 19.54 \\
         & sent & 11.89 \\
         & \textbf{cond - c} & \textbf{20.44} \\
         & cond & 18.32 \\
        \midrule
        \acs{GTE}  & sent - c & 7.16 \\
         & sent & 7.16 \\
         & \textbf{cond - c} & \textbf{20.40} \\
         & cond & 16.58 \\
        \midrule
        \acs{E5} & sent - c & 11.08 \\
         & sent & 3.77 \\
         & \textbf{cond - c} & \textbf{15.37} \\
         & cond & 6.07 \\
        \midrule
        \acs{SimCSE-large}  & $\operatorname{CONC}(s + c)$ & 5.59 \\
        & $\operatorname{CONC}(c + s)$ & 4.00 \\
        & $\operatorname{CONC}(s + c)$ - c & 8.32 \\
        & $\operatorname{\textbf{CONC}}\bm{(c + s)}$ \textbf{- c} & \textbf{8.58} \\
        \midrule
        \acs{SimCSE-base} & $\operatorname{CONC}(s + c)$ & 4.37 \\
        & $\operatorname{CONC}(c + s)$ & 1.25 \\
        & $\operatorname{\textbf{CONC}}\bm{(s + c)}$ \textbf{- c} & \textbf{7.05} \\
        & $\operatorname{CONC}(c + s)$ - c  & 6.00 \\
        \bottomrule
    \end{tabular}
    \caption{Spearman scores for different sentence embedding models and encoding settings.}
    \label{tab:sent_cond}
\end{table}

We evaluate model performance on different pooling methods, prompt settings, and sentence constructions. 
Moreover, we evaluate the supervised methods of linear and non-linear \acp{FFN}. 
Both the linear and non-linear \acp{FFN} are Siamese bi-encoders with weight-sharing. 
As explained in \autoref{sec:projection}, they take $f(c; I(s_1))$ and $f(c; I(s_2))$ as the input embeddings, and return the projected embeddings ${\rm CASE}(s_1,c)$ and  ${\rm CASE}(s_2,c)$ as the outputs, which can be compared using a similarity metric such as the cosine similarity.

Linear \ac{FFN} performs a linear transformation:
\begin{align}
    \label{eq:LP}
    \mathbf{z}=\mat{W}\vec{e}
\end{align}
where $\mat{W} \in \mathbb{R}^{d^{\prime} \times d}$ is the learned projection matrix.

Our non-linear \ac{FFN} is a one-layer \ac{FFN} with a activation function LeakyReLU:
\begin{align}
    \mathbf{z'} &= \operatorname{Dropout}\left(\operatorname{LeakyReLU}\left(\mat{W}' \vec{e}\right)\right)
    \label{eq:nonLP}
\end{align}
A dropout layer is applied to reduce any overfitting~\cite{hinton2012improvingneuralnetworkspreventing}.
Hyperparameters are tuned on a held-out validation set.
We select a learning rate of $10^{-3}$ and a batch size of 512.
The dropout rate is set to $15\%$ for non-linear \ac{FFN}.
\autoref{sec:app:ablation} provides the ablation study for the architectures of \acp{FFN}.

To address annotation errors in the original \ac{C-STS} dataset such as ambiguous and invalid conditions, \citet{L-CSTS} re-annotated the original validation set.
\citet{zhang-etal-2025-annotating} further identified the issues on both condition statements and ratings.
They improved the dataset by first revising the condition statements and then re-annotating the similarity ratings for both the training and validation sets.
To conduct a more accurate and reliable evaluation, we use the most recent revised and cleaned datasets by \citet{zhang-etal-2025-annotating}.\footnote{\url{https://huggingface.co/datasets/LivNLP/C-STS-Reannotated}}
We split their validation set 70\% vs. 30\% randomly to a validation set of 1983 instances and a test set of 851 instances.
In this C-STS dataset, there are 14176 unique instances with 4383 unique conditions.
As for each sentence pair ($s_1$, $s_2$), there are two conditions ($c_{\rm high}$ and $c_{\rm low}$).
There are 7088 unique sentence pairs, of which 5719 pairs have unique condition pairs ($c_{\rm high}$, $c_{\rm low}$).
In other words, 19.3\% of the sentence pairs share their conditions with other sentence pairs.

We use a single p3.24xl EC2 instance (8x V100 GPUs) for learning sentence embeddings, and a separate NVIDIA RTX A6000 GPU for supervised projection learning. 
Pytorch 2.0.1 with cuda 11.7 is used for \ac{FFN} projection. 
These settings are fixed across all experiments.
For reducing 4096-dimensional \ac{LLM}-based sentence embeddings to 512-dimensional, training a non-linear \ac{FFN} for \ac{CASE} takes less than 1 minute (wall-clock time).

\subsection{Evaluation Metrics}
\label{sec:metrics}

We evaluate the performance of a sentence encoder by the Spearman correlation coefficient between the cosine similarity scores computed using the embeddings produced by an encoder and the human similarity ratings on the test set.

\subsection{C-STS Measurement}
\label{sec:C-STS}

\begin{table}[t!]
\centering
\begin{tabular}{lcc}
\toprule
\textbf{Model} & \textbf{$I_{\text{iso}}$ (-c)} & \textbf{$I_{\text{iso}}$}\\
\midrule
\acs{NV} &  0.9611 & 0.9471\\
\acs{SFR} & 0.9463 & 0.9266\\
\acs{GTE} & 0.9490 & 0.9310\\
\acs{E5} &  0.8906 & 0.8368\\
\acs{SimCSE-base} & 0.9461 & 0.9297\\
\acs{SimCSE-large} &  0.9499 & 0.9406\\
\bottomrule
\end{tabular}
\caption{$I_{\text{iso}}$ values for each model. The left column of $I_{\text{iso}}$ (-c) lists values with the post-processing step of subtracting the embedding of condition \texttt{c}. $I_{\text{iso}}$ values close to 1 indicate high isotropy.}
\label{tab:approx_ipc_values_main}
\end{table}


To generate \ac{CASE}, we apply different ways to construct the prompt for \ac{LLM}-based embeddings and to concatenate the condition and sentence for \ac{MLM}-based embeddings. 
For \ac{LLM} embeddings, we have two main settings:
(a) $\operatorname{cond}=f(c; I(s))$, where we encode the condition given the sentence,
and (b) $\operatorname{sent}=f(s; I(c))$, where we encode the sentence given the condition as explained in \autoref{sec:LLM-embeddings}. 
For \ac{MLM} embeddings, we evaluate the two settings:
(a) $\operatorname{CONC}(c + s)$, where we concatenate sentence after the condition, and 
(b) $\operatorname{CONC}(s+c)$, where we concatenate condition after the sentence.
For each setting, we evaluate the effect of subtracting the condition embedding, $\vec{c}=f(c; I(\emptyset))$.

\begin{figure*}
    \centering
    \begin{subfigure}[b]{0.47\linewidth}
        \centering
        \includegraphics[width=\linewidth]{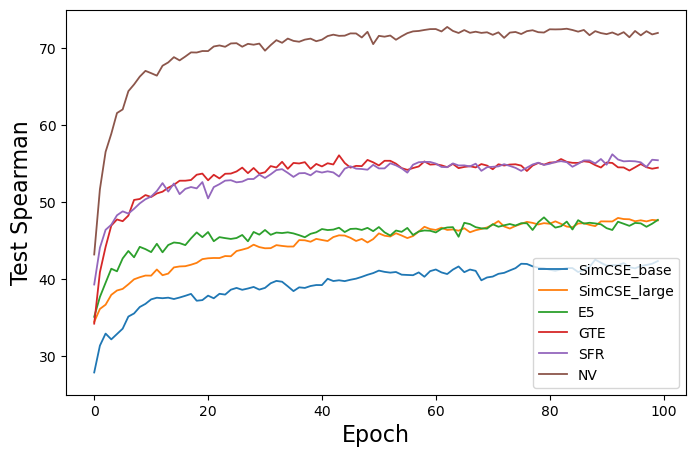}
        \caption{Test Spearman for linear \ac{FFN}}
        \label{fig:lp}
    \end{subfigure}
    \hfill
    \begin{subfigure}[b]{0.47\linewidth}
        \centering
        \includegraphics[width=\linewidth]{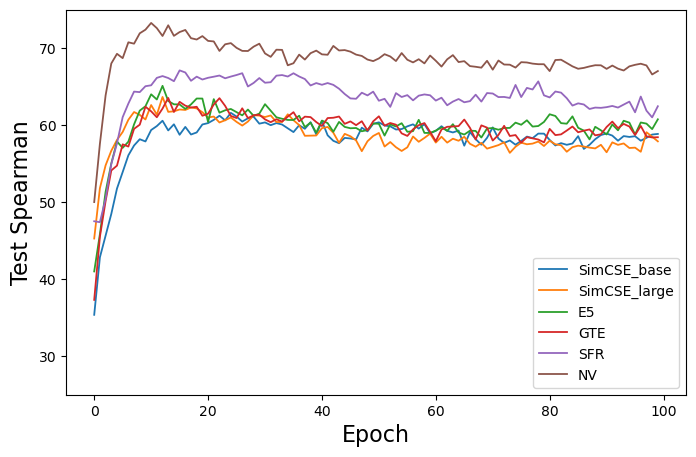}
        \caption{Test Spearman for non-linear \ac{FFN}}
        \label{fig:nonlp}
    \end{subfigure}
    \caption{Spearman correlation on test set for different models over training steps with dimensionality 512. 
    The $y$-axes of both subfigures are aligned, facilitating a direct comparison of the Spearman coefficients across the two line charts, with the same colour for the same model. Best viewed in colour.}
    \label{fig:spearman_epoch}
\end{figure*}

The test performance for different settings and models are shown in \autoref{tab:sent_cond}. 
For \ac{LLM} embeddings, \textbf{cond} consistently reports higher Spearman correlation than \textbf{sent}, suggesting that embedding the condition given the sentence is more effective for \ac{C-STS} measurement than embedding the sentence given the condition.
Moreover, we see that subtracting $\vec{c}$ further improves performance across all settings.
The former approach reduces the noise due to the tokens in a sentence, which are irrelevant to the given condition.
For \ac{MLM}-based embeddings, subtracting $\vec{c}$ also improves performance. 
This operation removes the sentence-independent components of the condition, thereby isolating the contrastive semantics — the aspects of the condition that are actually altered by the sentence. 
As a result, CASE focuses on the semantic variation between the two sentences under the given condition, rather than the redundant information shared across all condition phrases.

Additionally, we discovered that subtracting $\vec{c}$ in a post-processing step improves isotropy of the embeddings (full details in \autoref{sec:isotropy}).
We use the approximated isotropy metric $I_{\text{iso}}$ \cite{durdy2023metricsquantifyingisotropyhigh} as a numerically stable method to measure isotropy. 
$I_{\text{iso}}$ is estimated by randomly sampling unit vectors on the hypersphere and computing the ratio of their minimum to maximum alignment with the embeddings across random directions.
\autoref{tab:approx_ipc_values_main} shows that the post-processing step of subtracting the condition $\vec{c}$ gives higher $I_{\text{iso}}$ values, indicating the improvement of isotropy.
This makes subtle semantic differences between sentences under a given condition more distinguishable.
This is in-line with prior work reporting a positive correlation between isotropy and improved performance in embedding models \cite{rajaee-pilehvar-2022-isotropy, su2021whiteningsentencerepresentationsbetter}.
The effect of pooling method on performance is discussed in \autoref{sec:app:pooling}, where we find the \textbf{latent} pooling in \acs{NV} to perform best.
Therefore, we use the ($\operatorname{cond} - \vec{c}$) setting (which reports the best performance) for the six sentence encoders to conduct the subsequent experiments.

\begin{table}[!t]
    \centering
    \setlength{\tabcolsep}{4pt} 
    \begin{tabular}{lcc}
        \toprule
        Model & Non-linear \ac{FFN} & Linear \ac{FFN} \\
        \midrule
        \acs{NV} & \textbf{74.94} & 73.55 \\
        \acs{SFR} & \textbf{68.36} & 59.02 \\
        \acs{GTE} & \textbf{64.16} & 55.88 \\
        \acs{E5} & \textbf{65.07} & 48.25 \\
        \acs{SimCSE-large} & \textbf{63.64} & 48.15 \\
        \acs{SimCSE-base} & \textbf{61.52} & 42.22 \\
        \bottomrule
    \end{tabular}
    \caption{Spearman correlation of embedding models with supervised methods at reduced dimensionality $512$.}
    \label{tab:embedding_comparison}
\end{table}

\begin{figure}[!t]
    \centering
    \includegraphics[width=0.9\linewidth]{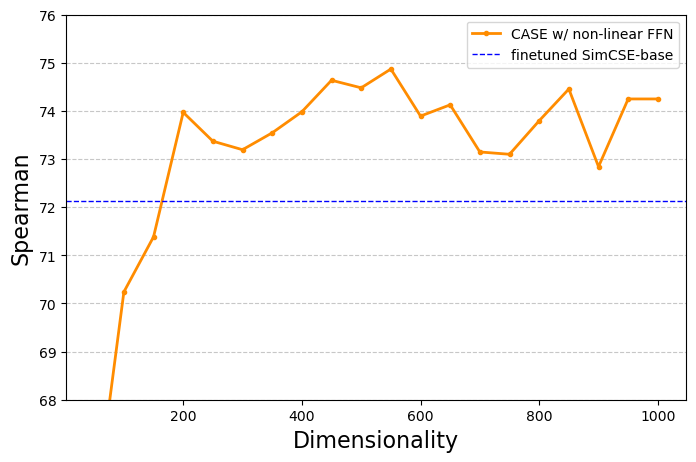}
    \caption{Spearman correlation coefficients of our CASE\ for NV embeddings on the test set over different dimensionalities. The horizontal dashed line represents the performance of fine-tuned SimCSE-base (768-dimensional).}
    \label{fig:r_dim}
\end{figure}

We show the training curves for our supervised \acp{FFN} in \autoref{fig:spearman_epoch}. 
Overall, the non-linear \acp{FFN} achieve significantly higher Spearman correlation than the linear \acp{FFN}.
Moreover, non-linear FFNs converge faster, typically reaching their peak performance within 20 epochs. 
The performance of the linear \acp{FFN}, gradually increases and eventually converges as the training progresses.
\acs{NV} consistently performs the best for both linear and non-linear \acp{FFN}. 


We compare sentence encoders in \autoref{tab:embedding_comparison} for a fixed (i.e. 512) dimensional projection from their original embeddings.
\acs{NV} obtains the highest Spearman coefficient across all sentence encoders and supervised methods.
\acs{GTE} and \acs{SFR} also achieve high Spearman coefficients under the non-linear \acp{FFN}.
\acs{SimCSE-large} and \acs{SimCSE-base} have the lowest Spearman coefficients.
Overall, we see that \ac{LLM}-based embeddings perform better than the \ac{MLM}-based embeddings on the \ac{C-STS} task.
Importantly, all embeddings are projected to the same lower dimensionality.

\begin{table}[t]
    \centering
    \resizebox{\columnwidth}{!}{\begin{tabular}{lccc}
        \textbf{Method} & \textbf{Spearman} & \textbf{dim} & \textbf{\#params}\\
        \toprule
        Deshpande et al. & & & \\
         ~w/ SimCSE-large     & 79.51 & 1024 & 354.3M\\
         ~w/ SimCSE-base & 72.12 & 768 & 124.1M\\
         CASE & & & \\
        ~w/ non-linear \ac{FFN}     & 74.94 & 512 & 2.1M\\
        \bottomrule
    \end{tabular}}
    \caption{Comparison of Spearman correlation, embedding dimensionality, and trainable parameters of previously proposed methods and our \ac{CASE}. \#params denotes the number of trainable parameters.}
    \label{tab:paper_comparison}
\end{table}

\begin{table*}[t!]
\centering
\footnotesize
\renewcommand{\arraystretch}{1}
\begin{tabular}{@{}p{\textwidth}@{}}
\toprule
\textbf{s1:} Young woman in \textcolor{orange}{orange} dress about to serve in \textcolor{blue}{tennis} game, on blue court with green sides. \\
\textbf{s2:} A girl playing \textcolor{blue}{tennis} wears a \textcolor{purple}{gray} uniform and holds her black racket behind her. \\
\textbf{cos(s1, s2)} \quad \ $0.5006 \rightarrow 0.9016$ \\[0.5ex]
\begin{tabularx}{\textwidth}{X|X}
\textbf{Condition 1:} color of the dress & \textbf{Condition 2:} name of the game \\
\textbf{Answer 1:} \textcolor{orange}{orange} \quad \textbf{Answer 2:} \textcolor{purple}{gray} \quad \textbf{Rating:} 1 
& \textbf{Answer 1:} \textcolor{blue}{tennis} \quad \textbf{Answer 2:} \textcolor{blue}{tennis} \quad \textbf{Rating:} 5 \\
\textbf{cos(s1, s2; c1)} \quad \ $0.4757 \rightarrow 0.2522$ & \textbf{cos(s1, s2; c2)} \quad \ $0.6233 \rightarrow 0.9660$ \\
\textbf{cos(s1, orange; c1)} \quad \ $0.1986 \rightarrow 0.3551$ & \textbf{cos(s1, tennis; c2)} \quad \ $0.1135 \rightarrow 0.6448$ \\
\textbf{cos(s2, gray; c1)} \quad \ $0.0559 \rightarrow 0.6061$ & \textbf{cos(s2, tennis; c2)} \quad \ $0.0983 \rightarrow 0.6426$ \\
\end{tabularx} \\
\midrule
\textbf{s1:} \textcolor{orange}{Two} snow \textcolor{blue}{skiers} with ski poles and snow skis, standing on top of a mountain with other skiers around them. \\
\textbf{s2:} \textcolor{purple}{A} \textcolor{blue}{skier} stands alone at the top of a snowy slope with blue skies and mountains in the distance. \\
\textbf{cos(s1, s2)} \quad \ $0.6318 \rightarrow 0.7702$ \\[0.5ex]
\begin{tabularx}{\textwidth}{X|X}
\textbf{Condition 1:} number of person & \textbf{Condition 2:} type of job \\
\textbf{Answer 1:} \textcolor{orange}{two} \quad \textbf{Answer 2:} \textcolor{purple}{one} \quad \textbf{Rating:} 1 
& \textbf{Answer 1:} \textcolor{blue}{skier} \quad \textbf{Answer 2:} \textcolor{blue}{skier} \quad \textbf{Rating:} 5 \\
\textbf{cos(s1, s2; c1)} \quad \ $0.6358 \rightarrow 0.2788$ & \textbf{cos(s1, s2; c2)} \quad \ $0.7502 \rightarrow 0.9539$ \\
\textbf{cos(s1, two; c1)} \quad \ $-0.0970 \rightarrow 0.7014$ & \textbf{cos(s1, skier; c2)} \quad \ $0.2488 \rightarrow 0.8016$ \\
\textbf{cos(s2, one; c1)} \quad \ $0.0227 \rightarrow 0.7953$ & \textbf{cos(s2, skier; c2)} \quad \ $0.3409 \rightarrow 0.8032$ \\
\end{tabularx} \\
\midrule
\textbf{s1:} A bunch of people standing around at the \textcolor{blue}{beach} with a \textcolor{orange}{kite} in the air. \\
\textbf{s2:} a \textcolor{blue}{beach} scene with a beach chair decorated with the Canadian Flag and surfers walking by with their \textcolor{purple}{surfboards} \\
\textbf{cos(s1, s2)} \quad \ $0.3988 \rightarrow 0.5350$ \\[0.5ex]
\begin{tabularx}{\textwidth}{X|X}
\textbf{Condition 1:} type of hobby & \textbf{Condition 2:} type of location \\
\textbf{Answer 1:} \textcolor{orange}{kite flying} \quad \textbf{Answer 2:} \textcolor{purple}{surf} \quad \textbf{Rating:} 1 
& \textbf{Answer 1:} \textcolor{blue}{beach} \quad \textbf{Answer 2:} \textcolor{blue}{beach} \quad \textbf{Rating:} 5 \\
\textbf{cos(s1, s2; c1)} \quad \ $0.4386 \rightarrow 0.4886$ & \textbf{cos(s1, s2; c2)} \quad \ $0.4825 \rightarrow 0.8582$ \\
\textbf{cos(s1, kite flying; c1)} \quad \ $0.3457 \rightarrow 0.7717$ & \textbf{cos(s1, beach; c2)} \quad \ $0.2254 \rightarrow 0.7281$ \\
\textbf{cos(s2, surf; c1)} \quad \ $0.1034 \rightarrow 0.6665$ & \textbf{cos(s2, beach; c2)} \quad \ $0.1129 \rightarrow 0.6703$ \\
\end{tabularx} \\
\bottomrule
\end{tabular}
\caption{Example of similarity scores for two conditions applied to the same sentence pair, based on non-linear \ac{FFN} with dimensionality $512$ on \ac{NV}. The table shows how supervised \ac{FFN} improves the \ac{CASE} for \ac{C-STS} task. cos$(\cdot, \cdot)$ denotes cosine similarity. Answer 1 and Answer 2 refer to the corresponding answers for Sentence 1 and Sentence 2 under conditions. The predicted similarity scores before and after applying supervised \ac{FFN} are listed on the left and right of the arrow. That is, the similarity score for original high-dimensional \ac{LLM}-based embeddings is on the left of the arrow, while the similarity score for \ac{CASE} is on the right.}
\label{tab:case_study}
\end{table*}

To study how the performance of the proposed \ac{CASE} varies with the dimensionality of the supervised projection, we plot the Spearman correlation measured on the test split of the \ac{C-STS} dataset in \autoref{fig:r_dim} using \acs{NV} embeddings.
Because we are specifically interested in examining how \ac{CASE} performs in the low-dimensional regime, which is attractive from a computational viewpoint, we consider dimensionalities in range [50, 1000].
Moreover, as a reference, we plot the performance of the previously proposed bi-encoder model by \cite{deshpande-etal-2023-c} for the SimCSE-base model, which is the smallest dimensional (768) encoder used in their work.
From \autoref{fig:r_dim} we see that for dimensionalities greater than 200, \ac{CASE} consistently outperforms SimCSE-base bi-encoder. 
This result has practical implications because it shows that we can reduce the high dimensionality of \acp{LLM}-based (\acs{NV}) embeddings from 4096 to 512 with an 8X compression, while preserving its high performance.
This result is consistent with Matryoshka Representation Learning (MRL)~\cite{kusupati2022matryoshka}, where they show that small-dimensional embeddings can maintain high performance in downstream tasks.
MRL is applicable with our \ac{MLM} and \ac{LLM} embeddings to further reduce the dimensionality,
while our \acp{FFN} project embeddings into a specific lower dimensionality rather than the nested granularities of MRL.
Our results show that embeddings can be efficiently compressed in the bi-encoder setting on the C-STS task.

We compare \ac{CASE} against the best bi-encoder models with fine-tuned SimCSE by \citet{deshpande-etal-2023-c}.
As shown in \autoref{tab:paper_comparison}, our proposed \ac{CASE} with non-linear \ac{FFN} achieves 74.9\% Spearman correlation, outperforming fine-tuned \acs{SimCSE-base} while using only 2.1M trainable parameters and a 512-dimensional embedding space. 
Compared with \acs{SimCSE-large}, \ac{CASE} maintains 94\% of its performance, but requires 50\% fewer dimensions and only 0.6\% of its trainable parameters. 
This highlights that \ac{CASE} is a simple but efficient method.

\subsection{Case Study}
\label{sec:interpretation}

To further illustrate how \ac{CASE} captures condition-dependent semantics, \autoref{tab:case_study} presents qualitative examples.
Similarity scores are computed and compared for three sentence pairs for the original high-dimensional \ac{LLM}-based embeddings and low-dimensional trained \ac{CASE}.
By comparing the similarity scores with and without the supervised projection, we can evaluate the ability of \ac{CASE} to focus on the condition-related information.
Compared to the unconditional similarity between the two sentences, conditional similarities scores computed using \acs{NV} embeddings align well with the human ratings.
For example, in the top row in \autoref{tab:case_study}, we see that the unconditional similarity between the two sentences reduces from $0.5006$ to $0.4747$ under $c_1$, while increasing to $0.6233$ under $c_2$.
Moreover, the conditional similarities are further appropriately amplified by \ac{CASE} using the supervised projection (i.e. decreasing to $0.2522$ under $c_1$, while increasing to $0.9660$ under $c_2$).
Specifically, \ac{CASE} reduces the similarity between the two sentences under the lower-rated condition, while increasing the same under the high-rated condition. 

We also treat the conditions as questions for sentences and extract the relevant information as answers to compare whether \ac{CASE} can focus on the condition-related information.
The overall similarity trend is consistent with the actual ratings.
For all sentence and answer pairs, the similarity scores increase after supervised projection. 
This demonstrates that \ac{CASE} can effectively capture the shift in conditional meaning between sentences under different conditions.

\begin{table}[t]
    \centering
    \begin{tabular}{l c c}
        \toprule
        \textbf{Method} & \textbf{Spearman} & \textbf{Dim} \\
        \midrule
        \multicolumn{3}{l}{\textit{Qwen3-8B (frozen)}} \\
        \hspace{3mm} w/o FFN & 24.59 & 4096 \\
        \hspace{3mm} w/ FFN (CASE) & 71.64 & \textbf{512} \\
        \midrule
        \multicolumn{3}{l}{\textit{Qwen3-8B (LoRA fine-tuned)}} \\
        \hspace{3mm} w/o FFN & \textbf{79.91} & 4096 \\
        \hspace{3mm} w/ FFN & 79.11 &  \textbf{512} \\
        \bottomrule
    \end{tabular}
    \caption{Comparison of fine-tuning. We report the Spearman correlation of the original and LoRA fine-tuned Qwen3-8B encoders, both with and without our supervised non-linear \ac{FFN}.}
    \label{tab:qwen_results}
\end{table}

\subsection{Comparision with Fine-tuning \ac{LLM} Encoders}
\label{sec:fine-tune}

We investigate the effects of fine-tuning \ac{LLM} encoders using LoRA~\cite{hu2022lora}, and compare its performance with our \ac{CASE} method.
Specifically, we use the high-performing encoder \textit{Qwen3-Embedding-8B} (\textbf{Qwen3-8B})\footnote{\url{https://huggingface.co/Qwen/Qwen3-Embedding-8B}} in the MTEB learderboard.
The encoder uses a 4096-dimensional embedding space with last-token pooling.
We fine-tune the Qwen3-8B on the \ac{C-STS} dataset with Pearson correlation loss computed over the cosine similarities.
For the LoRA configuration, we use a rank $r=32$, a scaling factor $\alpha=32$, and a dropout rate of $0.05$. 
Adaptation is applied to the query, key, value, and output projection layers of the attention mechanism. 
Fine-tuning Qwen3-8B requires over four hours of wall-clock time on two NVIDIA A100 (80GB) GPUs. 
In contrast, \ac{CASE} learns a supervised \ac{FFN} on pre-computed embeddings and completes training in under one minute on a single NVIDIA RTX A6000 GPU (approximately 10 seconds on an A100).
Moreover, LoRA fine-tuning of 8B-scale models requires substantial memory for model loading and activation storage, often necessitating at least 24 GB of GPU memory in practical settings, while \ac{CASE} supervision requires negligible GPU memory usage (<1 GB).

\autoref{tab:qwen_results} shows the results. 
Without fine-tuning, applying our \ac{CASE} method to the frozen Qwen3-8B yields a Spearman correlation 
of 71.64 after the supervised \ac{FFN}.
While fine-tuning the full encoder achieves a higher Spearman correlation of 79.91, it comes with substantial computational costs.
Our \ac{CASE} method achieves 90\% of the performance of the fine-tuned Qwen3-8B in a lightweight and resource-efficient setting.
Additionally, applying our supervised \ac{FFN} to the fine-tuned embeddings allows an 8$\times$ dimensionality reduction in the embeddings (from 4096 to 512), while retaining 99\% of the performance.

\section{Conclusion}
We propose \ac{CASE}, a method for computing the semantic textual similarity between two sentences under a given condition.
\ac{CASE} encodes the condition under a sentence, and then subtracts the embedding of the condition in a post-processing step.
Our experimental results show that \ac{LLM} embeddings consistently outperform \ac{MLM} embeddings for \ac{C-STS}.
Moreover, we introduce an efficient supervised projection learning method to improve the performance in \ac{C-STS} while projecting embeddings to lower-dimensional spaces for efficient similarity computations.

\section*{Acknowledgements}
Danushka Bollegala holds concurrent appointments as a Professor at the University of Liverpool and as an Amazon Scholar at Amazon.
This paper describes work performed at the University of Liverpool and is not associated with Amazon.

\section{Limitations}
\label{sec:limitations}


Our evaluations cover only English, which is a morphologically limited language.
To the best of our knowledge, \ac{C-STS} datasets have not been annotated for languages other than English, which has forced all prior work on \ac{C-STS} to conduct experiments using only English data.
However, the sentence encoders we used in our experiments support multiple languages.
Therefore, we consider it to be an important future research direction to create multilingual datasets for \ac{C-STS} and evaluate the effectiveness of our proposed method in multilingual settings.

There is a large number of sentence encoders (over 18,000 sentence encoders as in January 2026 evaluated in Hugging Face Hub\footnote{\url{https://huggingface.co/models}}).
However, due to computational limitations, we had to select a subset covering the best performing (top-ranked on MTEB leaderboard at the time of writing) models for our evaluations.

\section{Ethical Concerns}
\label{sec:ethics}

We did not collect or annotate any datasets in this project.
Instead, we use existing \ac{C-STS} datasets annotated and made available by \citet{deshpande-etal-2023-c}, \citet{L-CSTS}, and \citet{zhang-etal-2025-annotating}.
To the best of our knowledge, no ethical issues have been raised regarding those datasets.

We use multiple pre-trained and publicly available \ac{MLM}- and \ac{LLM}-based sentence encoders~\cite{Kaneko:2021}.
Both \acp{MLM} and \acp{LLM} are known to encode unfair social biases such as gender or racial biases.
We have not evaluated how such social biases would be influenced by the \ac{CASE} learning method proposed in this work.
Therefore, we consider it would be important to measure the social biases in \ac{CASE} created in this work before they are deployed in real-world applications.

\bibliography{csts.bib}

\appendix
\section*{Supplementary Materials}

\section{Models}
\label{sec:app:models}

To evaluate the effectiveness of the \ac{LLM}-based and \ac{MLM}-based sentence embeddings as described in \autoref{sec:CASE}, we apply six sentence encoders, out of which three are LLM-based: \textit{NV-Embed-v2} (4096 dimensional and uses latent pooling) (\textbf{\acs{NV}})\footnote{\url{https://huggingface.co/nvidia/NV-Embed-v2}}, \textit{SFR-Embedding-Mistral} (4096 dimensional and uses average pooling) (\textbf{\acs{SFR}})\footnote{\url{https://huggingface.co/Salesforce/SFR-Embedding-Mistral}}, 
\textit{gte-Qwen2-7B-instruct} (3584 dimensional and uses last token pooling) (\textbf{\acs{GTE}})\footnote{\url{https://huggingface.co/Alibaba-NLP/gte-Qwen2-7B-instruct}} and three are MLM-based: \textit{Multilingual-E5-large-instruct} (1024 dimensional and uses average pooling) (\textbf{\acs{E5}})\footnote{\url{https://huggingface.co/intfloat/multilingual-e5-large-instruct}}, \textit{sup-simcse-roberta-large} (1024 dimensional and uses last token pooling) (\textbf{\acs{SimCSE-large}})\footnote{\url{https://huggingface.co/princeton-nlp/sup-simcse-roberta-large}}, and \textit{sup-simcse-bert-base-uncased} (786 dimensional and uses last token pooling) (\textbf{\acs{SimCSE-base}})\footnote{\url{https://huggingface.co/princeton-nlp/sup-simcse-bert-base-uncased}}.

\section{Ablation Study for Supervised FFN}
\label{sec:app:ablation}

\autoref{tab:ablation} provides the ablation study of linear and non-linear \ac{FFN}, using \ac{NV} as a representative model (for other encoders, we observe similar trends).
We assess the impact of including a dropout layer.
For non-linear \ac{FFN}, we further compare the effects of different activation functions (ReLU, GeLU, SiLU, LeakyReLU) and the number of layers (one, two).
Note that based on preliminary experiments, we set the dropout rate to 20\% for the Linear FFN and 15\% for the Non-linear FFNs as the best configuration.
Additionally, for two-layer non-linear \acp{FFN}, the intermediate layer dimension is set to 1024, which performs best among the candidate set \{256, 768, 1024, 2048, 8192\}.

Linear \ac{FFN} without dropout gives higher performance than that with dropout, with a Spearman correlation of 73.55.
For non-linear \acp{FFN}, one-layer \acp{FFN} generally outperform two-layer \acp{FFN} by 1-3 points.
Additionally, introducing a dropout layer improves the performance, probably by reducing the impact of overfitting.
Among the four activation functions, LeakyReLU yields the best results for both two-layer and one-layer \acp{FFN}, achieving a Spearman correlation over 74.
Consequently, we proceed with the linear \ac{FFN} (without dropout) and the non-linear \ac{FFN} (one layer, activation function LeakyReLU, dropout of 15\%) for the subsequent experiments.

\begin{table}[t]
    \centering
    \small
    \begin{tabular}{lcc}
        \toprule
        \textbf{Activation} & \textbf{w/ Dropout} & w/o Dropout \\
        \midrule
        
        \multicolumn{3}{l}{\textbf{Linear FFN (4096$\to$512)}}\\ 
        & 72.90 & \textbf{73.55} \\
        \midrule
        
        \multicolumn{3}{l}{\textbf{Two-layer Non-linear FFN (4096$\to$1024$\to$512)}} \\
        
        \quad ReLU & 73.93 & 71.47 \\
        \quad GeLU & 67.66 & 69.06 \\
        \quad SiLU & 69.33 & 69.58 \\
        \quad LeakyReLU & 74.05 & 73.29 \\
        \midrule
        
        \multicolumn{3}{l}{\textbf{One-layer Non-linear FFN (4096$\to$512)}} \\

        \quad ReLU & 73.27 & 71.56 \\
        \quad GeLU & 71.05 & 70.56 \\
        \quad SiLU & 71.08 & 71.13 \\
        \quad LeakyReLU & \textbf{74.94} & 71.16 \\
                
        \bottomrule
    \end{tabular}
    \caption{Ablation study comparing Linear and Non-linear FFNs. Numbers in parentheses indicate layer dimensions.}
    \label{tab:ablation}
\end{table}

\section{Isotropy of Embeddings}
\label{sec:isotropy}

We first use the embedding-to-mean cosine similarity distribution to measure the isotropy of the embeddings. 
Given a set of embeddings $S = \{\vec{x}_1, \vec{x}_2, \dots, \vec{x}_n\}$, we first compute the mean embedding vector $\vec{\mu} = \frac{1}{n} \sum_{i=1}^{n} \vec{x}_i $. 
Then, for each embedding $ \vec{x}_i \in S $, we compute its cosine similarity with the mean vector $ \mu $, i.e., $\cos(\vec{x}_i, \vec{\mu}) = \frac{\vec{x}_i\T \vec{\mu}}{\norm{\vec{x}_i} \norm{\vec{\mu}}}$. 
The distribution of these embedding-to-mean cosine similarities is then analysed to characterise the embedding space -- a distribution sharply peaked near 1 indicates anisotropy, whereas a broader, more uniform distribution suggests a more isotropic geometry.

\begin{figure*}[h!]
    \centering
    \scriptsize

    \begin{subfigure}[b]{0.47\linewidth}
        \centering
        \includegraphics[width=\linewidth]{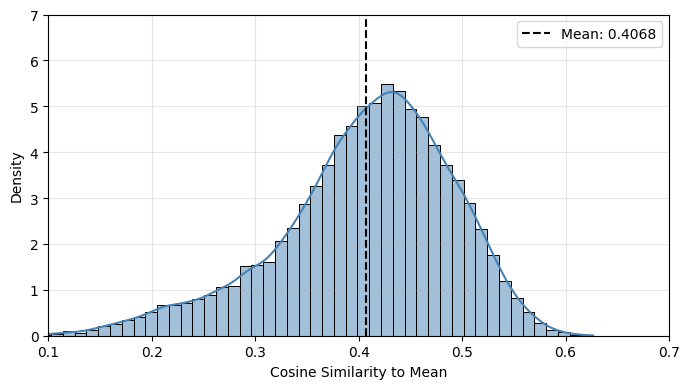}
        \caption{\textbf{\acs{NV}}: \texttt{cond - c}}
    \end{subfigure}
    \hfill
    \begin{subfigure}[b]{0.47\linewidth}
        \centering
        \includegraphics[width=\linewidth]{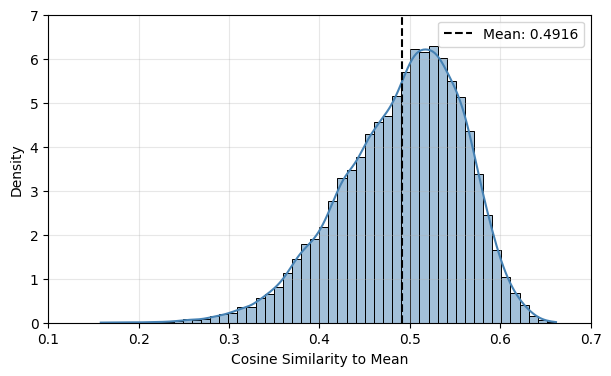}
        \caption{\textbf{\acs{NV}}: \texttt{cond}}
    \end{subfigure}

    \vspace{0.5em}

    \begin{subfigure}[b]{0.47\linewidth}
        \centering
        \includegraphics[width=\linewidth]{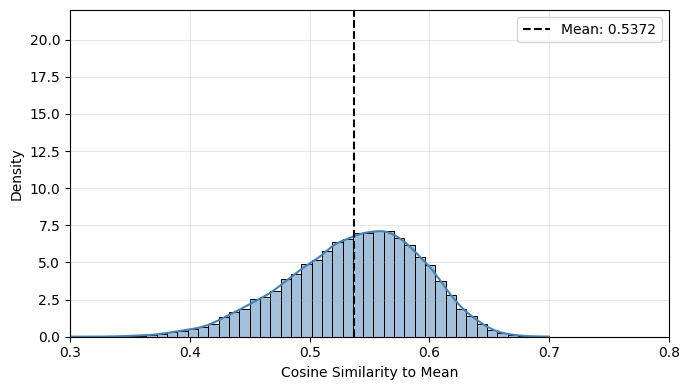}
        \caption{\textbf{\acs{SFR}}: \texttt{cond - c}}
    \end{subfigure}
    \hfill
    \begin{subfigure}[b]{0.47\linewidth}
        \centering
        \includegraphics[width=\linewidth]{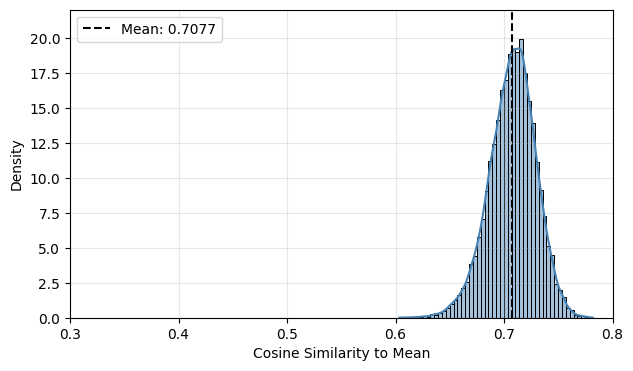}
        \caption{\textbf{\acs{SFR}}: \texttt{cond}}
    \end{subfigure}

    \vspace{0.5em}

    \begin{subfigure}[b]{0.47\linewidth}
        \centering
        \includegraphics[width=\linewidth]{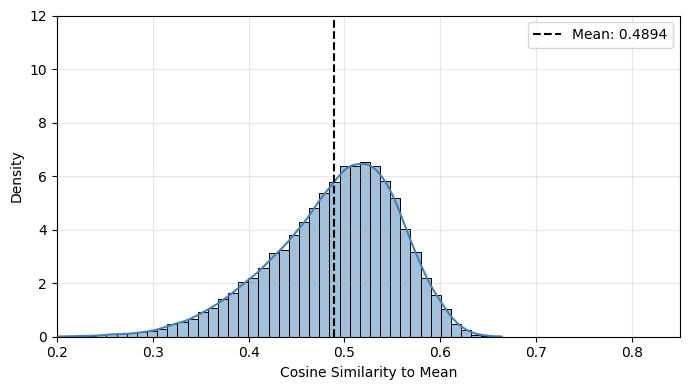}
        \caption{\textbf{\acs{GTE}}: \texttt{cond - c}}
    \end{subfigure}
    \hfill
    \begin{subfigure}[b]{0.47\linewidth}
        \centering
        \includegraphics[width=\linewidth]{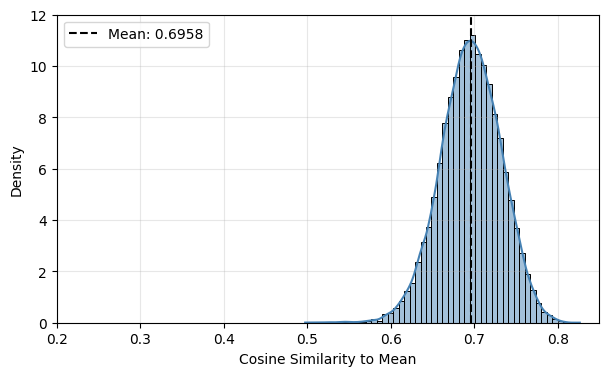}
        \caption{\textbf{\acs{GTE}}: \texttt{cond}}
    \end{subfigure}

    \caption{Embeddings-to-mean cosine similarity distributions across three \ac{LLM}-based models. Each row compares \texttt{cond - c} and \texttt{cond} representations.}
    \label{fig:isotropy_3models_llm}
\end{figure*}

\begin{figure*}[h!]
    \centering
    \scriptsize

    \begin{subfigure}[b]{0.47\linewidth}
        \centering
        \includegraphics[width=\linewidth]{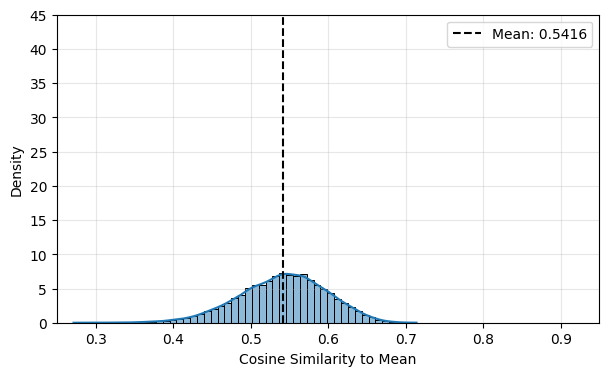}
        \caption{\textbf{\acs{E5}}: \texttt{cond - c}}
    \end{subfigure}
    \hfill
    \begin{subfigure}[b]{0.47\linewidth}
        \centering
        \includegraphics[width=\linewidth]{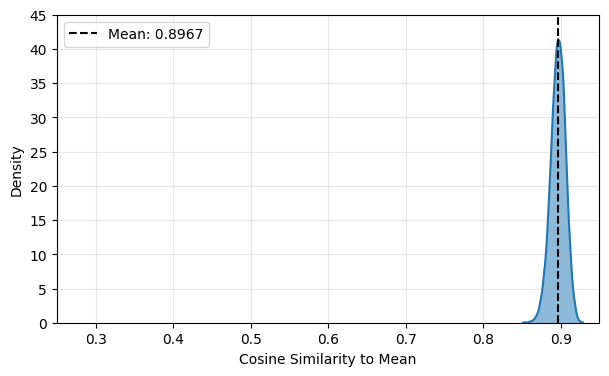}
        \caption{\textbf{\acs{E5}}: \texttt{cond}}
    \end{subfigure}

    \vspace{0.5em}

    \begin{subfigure}[b]{0.47\linewidth}
        \centering
        \includegraphics[width=\linewidth]{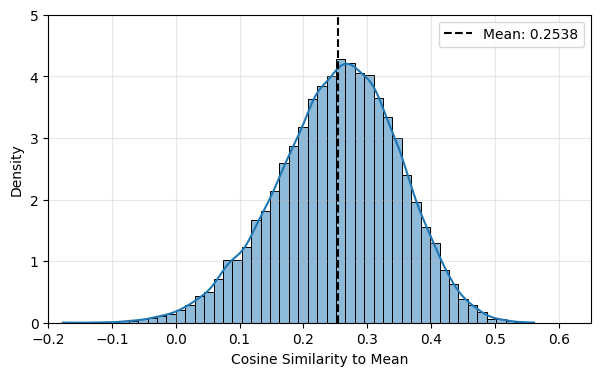}
        \caption{\textbf{\acs{SimCSE-base}}: $\operatorname{CONC}(c + s)$ - c}
    \end{subfigure}
    \hfill
    \begin{subfigure}[b]{0.47\linewidth}
        \centering
        \includegraphics[width=\linewidth]{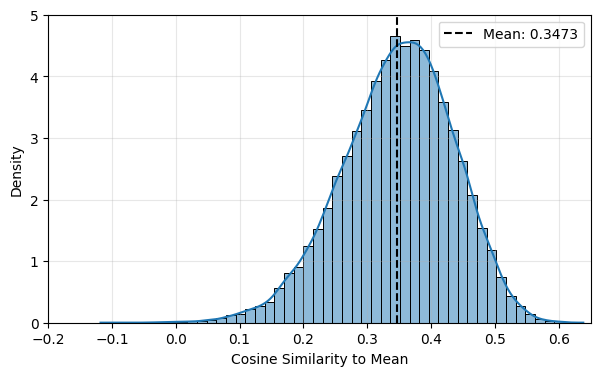}
        \caption{\textbf{\acs{SimCSE-base}}: $\operatorname{CONC}(c + s)$}
    \end{subfigure}

    \vspace{0.5em}

    \begin{subfigure}[b]{0.47\linewidth}
        \centering
        \includegraphics[width=\linewidth]{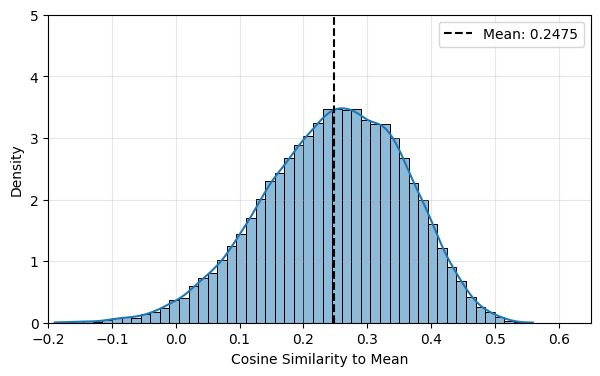}
        \caption{\textbf{\acs{SimCSE-large}}: $\operatorname{CONC}(c + s)$ - c}
    \end{subfigure}
    \hfill
    \begin{subfigure}[b]{0.47\linewidth}
        \centering
        \includegraphics[width=\linewidth]{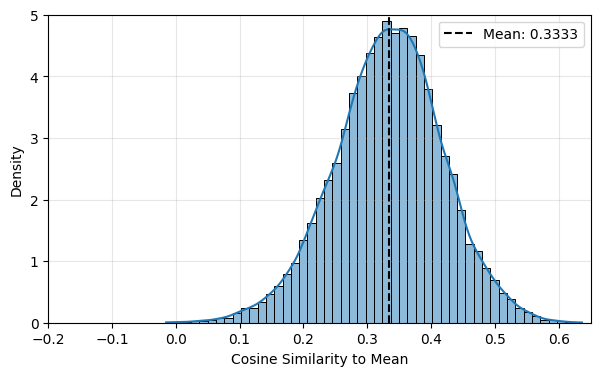}
        \caption{\textbf{\acs{SimCSE-large}}: $\operatorname{CONC}(c + s)$}
    \end{subfigure}

    \caption{Embeddings-to-mean cosine similarity distributions across three \ac{MLM}-based models. Each row compares two embedding types.}
    \label{fig:isotropy_3models_mlm}
\end{figure*}

From \autoref{app:tab:isotropy_nv}, \autoref{fig:isotropy_3models_llm}, and \autoref{fig:isotropy_3models_mlm}, we see that the embeddings after subtracting $\vec{c}$ have a lower mean cosine similarity to the mean vector and a higher standard deviation, indicating that they are more spread out in the embedding space.
In contrast, the embeddings without subtracting $\vec{c}$ are more clustered around a central direction (higher mean, lower standard deviation), reflecting anisotropy, a tendency for vectors to concentrate in a narrow region.
Therefore, embeddings after subtracting \texttt{c} tend to be more isotropic, indicating better distributional diversity.

\begin{table}[t!]
\centering
\resizebox{\columnwidth}{!}{ 
\begin{tabular}{llcc}
\toprule
\textbf{Model} & \textbf{Embedding Type} & \textbf{Mean} & \textbf{Std} \\
\midrule
\acs{NV} & \texttt{cond - c} & 0.407 & 0.084 \\
& \texttt{cond} & 0.492 & 0.067 \\
\midrule
\acs{SFR} & \texttt{cond - c} & 0.537 & 0.055 \\
& \texttt{cond} & 0.708 & 0.021 \\
\midrule
\acs{GTE} & \texttt{cond - c} & 0.489 & 0.067 \\
& \texttt{cond} & 0.696 & 0.036 \\
\midrule
\acs{E5} & \texttt{cond - c} & 0.542 & 0.056 \\
& \texttt{cond} & 0.897 & 0.010 \\
\midrule
\acs{SimCSE-base} & $\operatorname{CONC}(c + s)$ - c & 0.254 & 0.095 \\
& $\operatorname{CONC}(c + s)$ & 0.347 & 0.088 \\
\midrule
\acs{SimCSE-large} & $\operatorname{CONC}(c + s)$ - c & 0.248 & 0.110 \\
& $\operatorname{CONC}(c + s)$ & 0.333 & 0.085 \\
\bottomrule
\end{tabular}
}
\caption{Cosine similarity to mean vector: comparing mean and standard deviation of two embedding types across three \ac{LLM}-based and three \ac{MLM}-based models.}
\label{app:tab:isotropy_nv}
\end{table}

We use an efficient approximation method to compute the isotropy \cite{durdy2023metricsquantifyingisotropyhigh}.
Given a normalized embedding matrix $E \in \mathbb{R}^{n \times d}$ where each row represents a unit vector, instead of computing the eigenvectors of the covariance matrix, we randomly sample $k$ unit vectors $\vec{u}_1, \vec{u}_2, ..., \vec{u}_k$ from the unit hypersphere $\mathbb{S}^{d-1}$. 
For each sampled direction $\vec{u}_i$, we compute the function $F(\vec{u}_i) = \sum_{j=1}^{n} \exp(\vec{e}_j\T\vec{u}_i)$, where $\vec{e}_j$ represents the $j$-th embedding vector. 
The estimated isotropy is then defined as the ratio between the minimum and maximum values of $F$, given by \eqref{eq:IPC}.
\begin{align}
\label{eq:IPC}
    I_{\text{iso}} \approx \frac{\min_{i \in {1,...,k}} F(u_i)}{\max_{i \in {1,...,k}} F(u_i)}
\end{align}

$I_{\text{iso}}$ values close to 1 indicates that the embedding space shows high isotropy where vectors are uniformly distributed in all directions in the high-dimensional space instead of clustering in some small number of dominant directions. 
In contrast, when $I_{\text{iso}}$ approaches 0, it means a significant anisotropy in the embedding space. 
We use $k=1000$ random directions to compute the approximation of isotropy. 
\autoref{tab:approx_ipc_values} shows that the post-processing step of subtracting the condition \texttt{c} gives higher $I_{\text{iso}}$ values, indicating the improvement of isotropy.
Intuitively, when the embedding space is isotropic it becomes easier to differentiate between smaller similarity differences, thus improving the \ac{C-STS} estimation.

\begin{table}[t!]
\centering
\begin{tabular}{lcc}
\toprule
\textbf{Model} & \textbf{$I_{\text{iso}}$ (-c)} & \textbf{$I_{\text{iso}}$}\\
\midrule
\acs{NV} &  0.9611 & 0.9471\\
\acs{SFR} & 0.9463 & 0.9266\\
\acs{GTE} & 0.9490 & 0.9310\\
\acs{E5} &  0.8906 & 0.8368\\
\acs{SimCSE-base} & 0.9461 & 0.9297\\
\acs{SimCSE-large} &  0.9499 & 0.9406\\
\bottomrule
\end{tabular}
\caption{Isotropy values $I_{\text{iso}}$ for each model. The left column of $I_{\text{iso}}$ (-c) lists values with post-processing step of subtracting the condition \texttt{c}.}
\label{tab:approx_ipc_values}
\end{table}

\begin{table}[t!]
    \centering
    \footnotesize
    \renewcommand{\arraystretch}{1.1}
    \begin{tabular}{l c l c}
        \toprule
        \textbf{Model} & \textbf{Pooling} & \textbf{sent/cond?} & \textbf{Spear.} \\
        \midrule
        \acs{NV} & latent & sent - c & 16.98 \\
        & latent &  sent & 22.07 \\
        & \textbf{latent} &  \textbf{cond - c} & \textbf{37.61} \\
        & latent &  cond & 33.85 \\
        \midrule
        \acs{SFR} & last & sent - c & 11.88 \\
        &  last & sent & -0.18 \\
        &  last & cond - c & 19.28 \\
        &  last & cond & 13.00 \\
        &  average & sent - c & 19.54 \\
        &  average & sent & 11.89 \\
        &  \textbf{average} & \textbf{cond - c} & \textbf{20.44} \\
        &  average & cond & 18.32 \\
        \midrule
        \acs{GTE} & last & sent - c & 7.16 \\
        &  last & sent & 7.16 \\
        &  \textbf{last} & \textbf{cond - c} & \textbf{20.40} \\
        &  last & cond & 16.58 \\
        &  average & sent - c & 13.01 \\
        &  average & sent & 11.34 \\
        &  average & cond - c & 17.87 \\
        &  average & cond & 18.42 \\
        \midrule
        \acs{E5} & average & sent - c & 11.08 \\
        &  average & sent & 3.77 \\
        &  \textbf{average} & \textbf{cond - c} & \textbf{15.37} \\
        &  average & cond & 6.07 \\
        &  last & sent - c & 9.28 \\
        &  last & sent & 0.49 \\
        &  last & cond - c & 8.90 \\
        &  last & cond & 3.05 \\
        \midrule
        \multicolumn{2}{l}{\acs{SimCSE-large}}  & $\operatorname{CONC}(s + c)$ & 5.59 \\
        &  & $\operatorname{CONC}(c + s)$ & 4.00 \\
        &  & $\operatorname{CONC}(s + c)$ - c & 8.32 \\
        &  & $\operatorname{\textbf{CONC}}\bm{(c + s)}$ \textbf{- c} & \textbf{8.58} \\
        \midrule
        \multicolumn{2}{l}{\acs{SimCSE-base}} & $\operatorname{CONC}(s + c)$ & 4.37 \\
        &  & $\operatorname{CONC}(c + s)$ & 1.25 \\
        &  & $\operatorname{\textbf{CONC}}\bm{(s + c)}$ \textbf{- c} & \textbf{7.05} \\
        &  & $\operatorname{CONC}(c + s)$ - c  & 6.00 \\
        \bottomrule
    \end{tabular}
    \caption{Spearman scores for sentence embedding models. The original dimensionality of each model is indicated in parentheses. \textbf{latent} denotes latent attention pooling for \ac{NV}, whereas \textbf{last} and \textbf{average} correspond to last token pooling and average pooling, respectively.}
    \label{app:tab:sent_cond_full}
\end{table}
\section{Full Results for all Models}
\label{sec:app:pooling}

Each \ac{LLM} encoder has its own optimal pooling method, recommended by the original authors of those models.
From \autoref{app:tab:sent_cond_full} we see that the performance varies significantly depending on the pooling method being used, while the latent attention pooling used in \textbf{NV} reporting the best results.
Note that NV does not support \textbf{last} or \textbf{average} pooling, while \textbf{latent} pooling is not supported by the other models.

\end{document}